\documentclass[letterpaper]{article} 
\usepackage[preprint]{aaai2027}  
\usepackage[hyphens]{url}  
\usepackage{graphicx} 
\usepackage{natbib}  
\usepackage{caption} 
\usepackage{algorithm}
\usepackage{algorithmic}

\usepackage{amsfonts}  
\usepackage{amssymb}
\usepackage{amsmath}
\usepackage{xcolor}
\usepackage{multirow}
\usepackage{colortbl}
\usepackage{dsfont}

\usepackage{mathtools}
\usepackage{amsthm}
\newtheorem{theorem}{Theorem} 
\usepackage{newfloat}
\usepackage{listings}
\DeclareCaptionStyle{ruled}{labelfont=normalfont,labelsep=colon,strut=off} 
\floatstyle{ruled}
\newfloat{listing}{tb}{lst}{}
\floatname{listing}{Listing}

\usepackage{booktabs}

\newcommand{\methodname}{PlanPO}

\title{\methodname{}: Group Planning-Aware Policy Optimization for Multi-Turn Agentic LLMs}
\author{
    Dayang Liang\textsuperscript{\rm 1},
    Liyuan He\textsuperscript{\rm 2},
    Xuan Feng\textsuperscript{\rm 3},
    Shuxin Li\textsuperscript{\rm 4},
    Bo An\textsuperscript{\rm 4},
    Yunlong Liu\textsuperscript{\rm 1}\corresponding
}
\affiliations{
    \textsuperscript{\rm 1}Department of Automation, Xiamen University, Xiamen, China \\
    \textsuperscript{\rm 2}School of Artificial Intelligence, Shanghai Jiaotong University, Shanghai, China \\
    \textsuperscript{\rm 3}College of Cyberspace Security, Jinan University, Guangzhou, China \\
    \textsuperscript{\rm 4}College of Computing and Data Science, Nanyang Technological University, Singapore \\

    Corresponding Email: ylliu@xmu.edu.cn
}

\begin{document}

\maketitle

\begin{abstract}
Group-relative policy optimization has emerged as a key paradigm for training agentic large language models (LLMs) on multi-turn interactive tasks. However, most existing variants fail to distinguish advantages among successful trajectories even when these trajectories differ substantially in their interaction efficiency. For instance, circuitous successes are often assigned the identical outcome reward, causing advantage collapse and severe performance bottlenecks. To this end, we propose \textit{Group Planning-aware Policy Optimization \textbf{(\methodname{})}}, a simple yet effective RL method for learning generalizable planning abilities beyond task-specific high-quality behavior patterns. Specifically, \methodname{} introduces coarse-to-fine advantage signals, which capture the relative differences in \textit{trajectory-level} lengths and \textit{turn-level} response lengths \textit{conditioned on successful trajectories} sampled for the same task. Within the group-relative optimization structure, this enables agents to actively learn generalizable and deliberate behaviors spanning interaction planning and textual generation from high-quality rollouts, without degenerating into vanilla length minimization. Experimentally, \methodname{} improves over GRPO by \textbf{27.2\%} on average across the challenging multi-turn benchmarks ALFWorld, WebShop, and SciWorld, outperforming recent powerful baselines while incurring negligible additional training cost.
\end{abstract}

\section{Introduction}

Large Language Models (LLMs) have demonstrated remarkable progress across a wide range of complex multi-turn tasks, including information retrieval \cite{liao2025agentmaster,eugeneformact}, web navigation \cite{rawles2025androidworld}\cite{zhang2026prune4web}, code generation \cite{zhang2024codeagent,dai2026group}, and embodied interaction \cite{li2024embodied,zhang2026embodied}. 
Recent work has increasingly explored agentic reinforcement learning (RL) with verifiable outcome rewards \cite{zhao2025learning,dai2026group,li2026relook}, particularly Group Relative Policy Optimization (GRPO) \cite{shao2024deepseekmath}, to fine-tune open-source LLMs such as Qwen2.5 \cite{qwen2.5}, thereby improving the capabilities of LLM agents in multi-turn tasks.

However, a central challenge in on-policy agentic RL lies in \textit{enriching the reward signals of rollout trajectories while improving data utilization}  \cite{zhang2025landscape,zhang2025count,wang2026information,penghiper}. Many early studies addressed this issue by introducing value models, such as critics \cite{schulman2017proximal,dai2025cde} and process reward models (PRMs) \cite{chae2026web}, to evaluate turn-level behaviors. Yet these models may introduce estimation or proxy bias, and incur substantial memory costs \cite{chae2026web}. Recent group-optimization approaches instead mine informative rollouts to craft turn-level reward signals and achieve discriminative advantages. For example, HiPER \cite{penghiper} computes the multiple returns of trajectory segments via sub-task decomposition, while GiGPO \cite{feng2026group} constructs step advantage signals over the action space by identifying repeated anchor states. Other methods, such as R3L \cite{shi2026r} and GVPO \cite{zhang2025gvpo}, establish turn-level credit from failure reflection and diverse code execution feedback. However, most methods rely on labor-intensive manual design and empirical heuristics, which limits their generality across task settings. As a result, advantage collapse within successful rollout groups remains difficult to mitigate in a broadly applicable way \cite{zhao2026aem}.

\begin{figure}[t]
   \begin{center}
   \includegraphics[width=1.0\linewidth]{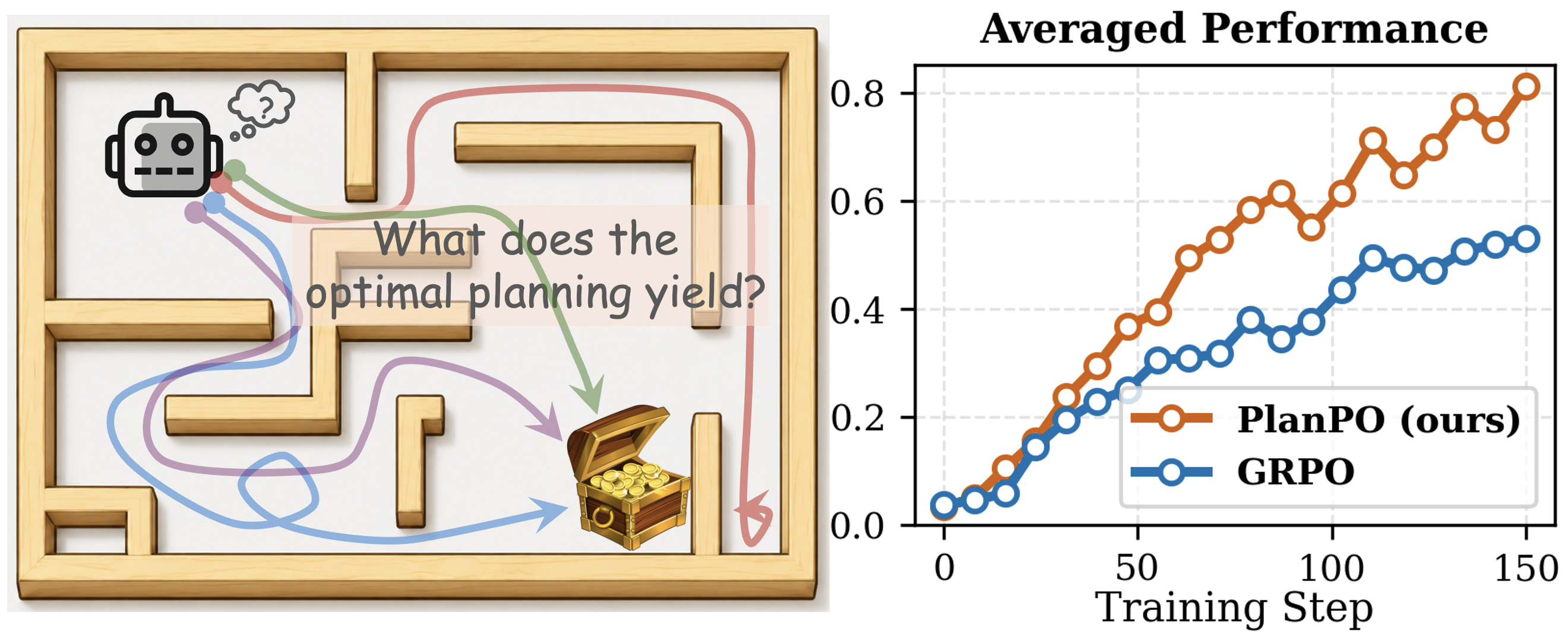}
   \end{center}
   \vspace{-0.1in}
   \caption{\textbf{Left:} Successful rollouts are not equally informative. Rollouts that reach the same task goal, can differ substantially in length, directness, and reasoning quality, while the most optimal strategies may reveal more sophisticated and generalizable capabilities. \textbf{Right:} Averaged  normalization performance comparison across ALFWorld, WebShop, and SciWorld environments using the Qwen2.5-1.5b model.}
   \label{fig:abs_intro}
   \vspace{-10pt}
\end{figure}

\textit{This motivates the question of how to enrich rollout-driven training signals for long-horizon tasks in a simple, effective, and more task-general manner.} We begin by revisiting a naturally available yet underutilized signal in group rollouts, i.e., the length profiles across both turn-level interaction trajectories and token-level generated responses. Figure \ref{fig:abs_intro} provides an abstract illustration of this intuition. Specifically, our key observation is that many inefficiencies in agentic RL manifest as excessive interaction or generation length. In multi-turn interactions, agents may hesitate between states, repeatedly visit similar observations, or enter dead ends before eventually completing the task. A similar issue arises in token-level textual responses. Given the same question or turn-level observation, sampled responses may produce correct actions while still containing unnecessarily verbose, convoluted, or even logically flawed reasoning traces. Nevertheless, both inefficient turns and reasoning tokens can still share identical success rewards just like the superior solutions. Crucially, treating such heterogeneous successes as equally preferable weakens distinguishable signals and the underlying abilities, while allowing noisy rollouts to degrade training quality and impose substantial performance bottlenecks.

To address this, we present \textit{Group Planning-aware Policy Optimization (\methodname{})}, an effective group-based RL method for learning generalizable planning abilities beyond specific planful behaviors. 
Specifically, within the set of successful trajectories sampled in same task, \methodname{} normalizes outcome rewards by turn-level trajectory lengths and token-level response lengths, constructing coarse-to-fine dense reward signals. Rather than directly summing these dense rewards, \methodname{} then computes their relative advantages separately and combines them through a weighted formulation, preserving discriminative information at different granularities \cite{liu2026gdpo}. Additionally, unlike generic length-based reward shaping \cite{pardo2018time,liu2025learn}, \methodname{} performs multiscale length normalization only conditioned on successes with group-relative structure. Thus, successful completion is a prerequisite for group length normalization. Empirically, we show that these success-conditioned relative advantages help agents acquire planning behaviors that generalize beyond specific high-quality trajectories, rather than merely imitating task-specific success patterns.


We evaluate \methodname{} on three challenging multi-turn benchmarks, ALFWorld \cite{shridhar2020alfworld}, WebShop \cite{yao2022webshop}, and SciWorld \cite{wang2022scienceworld}, using Qwen2.5-1.5B-Instruct and Qwen2.5-7B-Instruct. \methodname{} consistently outperforms recent strong baselines while incurring negligible additional training cost, specifically improving over GRPO by \textbf{27.2\%} on average and notably achieving a \textbf{24.3\%} gain on out-of-distribution tasks in ALFWorld. Extensive ablations and analyses further validate its effectiveness and generalizability. Our contributions are threefold.

\begin{itemize}
\item We identify successful-rollout heterogeneity as a critical bottleneck in group-relative optimization, where redundant turns and reasoning can mask rollout quality and then limit policy learning.

\item We propose \methodname{}, which constructs success-conditioned length-normalized advantages, encouraging agents to learning planning-aware abilities beyond specific behavioral patterns.

\item We show that \methodname{} consistently outperforms recent strong baselines across ALFWorld, WebShop, and SciWorld, while incurring negligible additional training cost.
\end{itemize}

\section{Preliminaries}

\paragraph{Multi-Turn Agentic RL.}

We consider RL for LLM-based agents that interact with an environment over multiple turns, where the interaction process is formulated as a finite-horizon Markov Decision Process (MDP). Given a task instance $\boldsymbol{x}\in p(X)$, at each turn $t=1,2,\ldots,T$, an agent policy $\pi_\theta$ observes a state $\boldsymbol{s}_t\in \mathcal{S}$ and generates a textual action $\boldsymbol{a}_t\in \mathcal{A}$, and then transitions to the next state $\boldsymbol{s}_{t+1}\in \mathcal{S}$ while yielding a scalar reward $r_t \in \mathbb{R}$. The interaction unfolds as a trajectory $\boldsymbol{\tau}=\{(\boldsymbol{s}_1,\boldsymbol{a}_1,r_1),(\boldsymbol{s}_2,\boldsymbol{a}_2,r_2),...,(\boldsymbol{s}_T,\boldsymbol{a}_T,r_T)\}$, where $T$ denotes the trajectory length in interaction turns. In practical tasks such as ALFWorld, the agent generates a response as its action for each observation. This response is structured within \texttt{<think></think>} and \texttt{<action></action>} tags, where the former contains the reasoning process, while the latter specifies the actual action executed in the environment. Notably, for most task settings, reward signals are sparse and delayed, e.g., the environment provides an outcome reward $R(\boldsymbol{\tau})$ only after the trajectory terminates. 

\subsection{Group-relative Policy Optimization}

Recent agentic RL methods for LLMs commonly adopt a group-relative policy optimization paradigm. Given a task instance $\boldsymbol{x}$, the old policy $\pi_{\theta_{\rm old}}$ samples a group of $N$ candidate trajectories $\mathcal{G}_x=\{\boldsymbol{\tau}_1,\boldsymbol{\tau}_2,\ldots,\boldsymbol{\tau}_N\}$, where each trajectory corresponds to one complete rollout. Each trajectory $\boldsymbol{\tau}_i$ receives a scalar reward $R(\boldsymbol{\tau}_i)$ that reflects the overall quality or success of the generated outcome. Instead of learning an advantage function $A(\boldsymbol{s}_t,\boldsymbol{a}_t)$ with critic networks in PPO \cite{schulman2017proximal}, group-based RL computes the advantage using only statistics within the sampled group:
\[
A(\boldsymbol{\tau}_i)
=
\mathtt{GroupNormalization}\left(\{R(\boldsymbol{\tau}_i)\}_{i=1}^{N}\right).
\]
In GRPO \cite{shao2024deepseekmath}, the advantage is evaluated by normalizing each trajectory reward with the mean and variance of group rewards $(\{R(\boldsymbol{\tau}_i)\}_{i=1}^{N})$. This sampling-based estimator reduces the memory and computational overhead introduced by the critic architecture in conventional PPO. 

\begin{figure*}[t]
    \centering
    \includegraphics[width=\linewidth]{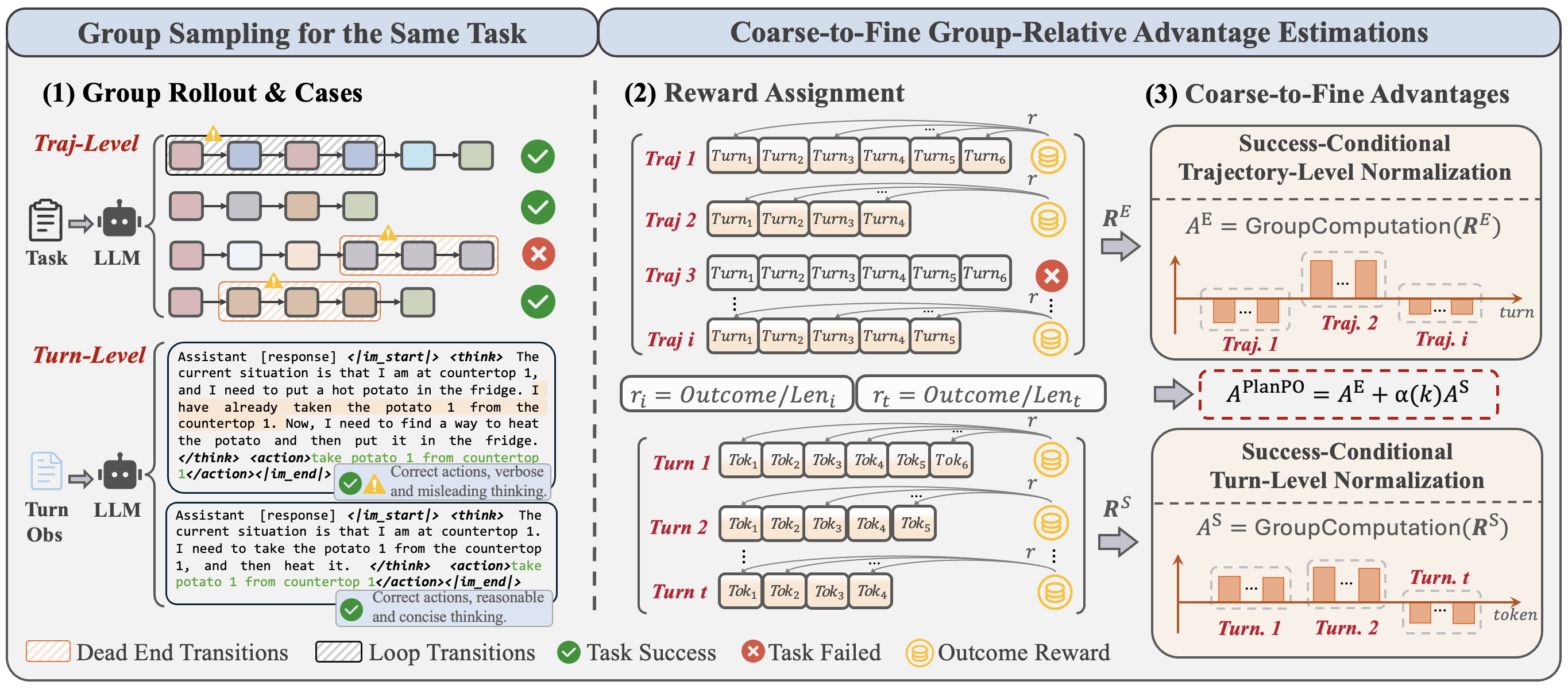}
    \caption{\textbf{Overview of \methodname{}.}
    \textbf{Left}: Rollouts sampled for the same task can reach the same outcome through interaction paths and textual responses of markedly different quality.
    \textbf{Middle}: \methodname{} converts the outcome into success-conditioned trajectory- and turn-level scores using trajectory and response lengths, respectively.
    \textbf{Right}: The two scores are normalized separately within the same-task group and combined into a coarse-to-fine advantage for policy optimization.}
    \label{fig:gppo}
    \vspace{-10pt}
\end{figure*}

\section{Group Planning-Aware Policy Optimization for Multi-turn Agentic LLMs}
\label{sec:method}
We propose \textit{group-relative Planning-aware Policy Optimization (\methodname{})}, a simple and effective group-relative RL method for learning high-level planning strategies beyond task-specific high-quality rollouts. We begin with our motivation, followed by the introduction of the coarse-to-fine advantage design, and conclude with the policy optimization loss and theoretical analysis.

\subsection{Motivation}
\label{sec:motivation}
As illustrated in the right of Figure~\ref{fig:gppo}, given the same task, some rollouts reach the goal through short and coherent interaction paths, whereas others involve redundant state transitions or even enter loops and dead ends. At the turn level, verbose reasoning can also introduce inconsistencies or misleading intermediate claims. For example, as illustrated, given a task or turn observation, although the final sampled actions "\texttt{<action>take potato 1 from countertop 1</action>}" are all correct, the thought process "\texttt{I have already taken the potato ...}" treats it as already happened. Such cases are common in the rollouts yet introduce hallucinated or logically inconsistent reasoning traces.

\textit{How should we quantify the quality of successful rollouts?} Existing solutions may introduce post-hoc reflection~\cite{shi2026r} or agentic verifier models \cite{zhang2026agentv}, but these signals are often costly and heuristic. The aforementioned observations motivate us to leverage multiscale rollout length as an initial signal. However, unconditional length-based reward shaping would degenerate into meaningless length minimization, even distorting the representation space of LLMs. Our goal is instead to comparatively learn planful behaviors within successful rollouts, while promoting task-general planning capabilities rather than fitting task-specific patterns. This objective naturally aligns with group-relative policy optimization. Below, we progressively construct the above conditional advantage signals.

\subsection{Trajectory Length-Normalized Advantage}
\label{sec:trajectory_advantage}

To achieve this, we first instantiate the success-conditioned length signal at the trajectory (or episode) level. For successful rollouts with the same outcome reward, trajectory length serves as a coarse proxy for planning efficiency across environment interactions. We therefore normalize the outcome reward by trajectory length only for successful rollouts, and compute the conditional group-relative advantage.

Formally, given a task instance $\boldsymbol{x}$, the policy $\pi_{\theta_{\rm old}}$ samples $N$ trajectories
$\{\boldsymbol{\tau}_1,\boldsymbol{\tau}_2,\ldots,\boldsymbol{\tau}_N\}$ start from the identical and initial state $\boldsymbol{s}_0$. Each rollout trajectory is represented as $\boldsymbol{\tau}_i=\{(\boldsymbol{s}_{i,1},\boldsymbol{a}_{i,1},r_{i,1}),(\boldsymbol{s}_{i,2},\boldsymbol{a}_{i,2},r_{i,2}),...,(\boldsymbol{s}_{i,T_i},\boldsymbol{a}_{i,T_i},r_{i,T_i})\}$, where $T_i$ represents the number of turns in $i$-th trajectory. In our task setting, each trajectory receives only a terminal outcome reward $R(\boldsymbol{\tau}_i)=10$ when the task goal is reached. We then denote the trajectory-level group of sampled trajectories and rewards as:
\begin{equation}
\mathcal{G}_{\boldsymbol{x}}^{\mathrm{E}}
=
\left\{
\bigl(\boldsymbol{\tau}_1,R(\boldsymbol{\tau}_1)\bigr),
\bigl(\boldsymbol{\tau}_2,R(\boldsymbol{\tau}_2)\bigr),
\ldots,
\bigl(\boldsymbol{\tau}_N,R(\boldsymbol{\tau}_N)\bigr)
\right\},
\label{eq:episode_group}
\end{equation}
where the superscript $\mathrm{E}$ denotes the episode, i.e., trajectory level. Let $\mathcal{G}_{\boldsymbol{x}}^{\mathrm{U}}\subseteq\mathcal{G}_{\boldsymbol{x}}^{\mathrm{E}}$ denote the subset of successful trajectories, and let $\mathds{1}[\boldsymbol{\tau}_i\in\mathcal{G}_{\boldsymbol{x}}^{\mathrm{U}}]$ indicate whether $\boldsymbol{\tau}_i$ succeeds. For each trajectory, we define its length-normalized episode reward as, 
\begin{equation}
R^{\mathrm{E}}(\boldsymbol{\tau}_i)=\mathds{1}[\boldsymbol{\tau}_i\in\mathcal{G}_{\boldsymbol{x}}^{\mathrm{U}}]\,R(\boldsymbol{\tau}_i)/T_i,
\end{equation}
where $T_i$ is the number of valid interaction turns in $\boldsymbol{\tau}_i$. The corresponding trajectory-level advantage $A^{\mathrm{E}}$ is computed by group-relative normalization:
\begin{equation}
A^{\mathrm{E}}(\boldsymbol{\tau}_i)
=
\frac{
R^{\mathrm{E}}(\boldsymbol{\tau}_i)
-
\operatorname{mean}
\left(
\left\{
R^{\mathrm{E}}(\boldsymbol{\tau}_j)
\right\}_{j=1}^{N}
\right)
}{
F_{\mathrm{norm}}
\left(
\left\{
R^{\mathrm{E}}(\boldsymbol{\tau}_j)
\right\}_{j=1}^{N}
\right)
}.
\label{eq:episode_advantage}
\end{equation}
Here, $F_{\mathrm{norm}}(\cdot)$ is the normalization factor, instantiated as either $\operatorname{std}(\cdot)+\epsilon$ or $1$ \cite{feng2026group}, with the former usually used by default, where $\epsilon$ is a small constant for numerical stability \cite{shao2024deepseekmath}. The trajectory-level $A^{\mathrm{E}}(\boldsymbol{\tau}_i)$ is broadcast to each turn in the $i$-th trajectory, providing a trajectory-wide discrimination signal among successful rollout trajectories that complete the same task but differ in process quality or long-term planning.

\subsection{Response Length-Normalized Advantage}
\label{sec:response_advantage}
While the above advantage provides learning signals for discriminating trajectories, each response turn within a trajectory still lacks fine-grained credit. Next, we compute group-relative advantages only among successful responses, so that the agent learns that planful responses are desirable only when the corresponding actions remain correct. Additionally, we flatten all successful responses generated for the same task into one group for advantage computation. 

We then employ a similar group relative advantage structure to achieve the above idea. Formally, for the $t$-th interaction turn in trajectory $\boldsymbol{\tau}_i$, we collect all active responses into a turn-level group,
\begin{equation}
\mathcal{G}_{\boldsymbol{x},t}^{\mathrm{S}}
=
\left\{
\bigl(\boldsymbol{a}_{i,t}, R(\boldsymbol{\tau}_i)\bigr)
\mid
\boldsymbol{\tau}_i\in\mathcal{G}_{\boldsymbol{x}}^{\mathrm{E}},\ t\leq T_i
\right\},
\label{eq:response_group}
\end{equation}
where $\mathrm{S}$ denotes the step, i.e., turn level, and $L_{i,t}=|\boldsymbol{a}_{i,t}|$ is the token length of the response generated by trajectory $\boldsymbol{\tau}_i$ at turn $t$. Similar to the trajectory-level case, we define the response length-normalized reward as,
\begin{equation}
R^{\mathrm{S}}(\boldsymbol{a}_{i,t})=\mathds{1}[\boldsymbol{\tau}_i\in\mathcal{G}_{\boldsymbol{x}}^{\mathrm{U}}]R(\boldsymbol{\tau}_i)/L_{i,t}.
\end{equation}
The turn-level relative advantage is then computed within the active response group:
\begin{equation}
A^{\mathrm{S}}(\boldsymbol{a}_{i,t})
=
\frac{
R^{\mathrm{S}}(\boldsymbol{a}_{i,t})
-
\operatorname{mean}
\left(
\left\{
R^{\mathrm{S}}(\boldsymbol{a}_{j,t})
\mid
\boldsymbol{a}_{j,t}\in\mathcal{G}_{\boldsymbol{x},t}^{S}
\right\}
\right)
}{
F_{\mathrm{norm}}
\left(
\left\{
R^{\mathrm{S}}(\boldsymbol{a}_{j,t})
\mid
\boldsymbol{a}_{j,t}\in\mathcal{G}_{\boldsymbol{x},t}^{S}
\right\}
\right)
}.
\label{eq:response_advantage}
\end{equation}
Here, the resulting turn advantage $A^{\mathrm{S}}(\boldsymbol{a}_{i,t})$ is assigned to all tokens in the response $\boldsymbol{a}_{i,t}$, which provides a fine-grained supervision signal for successful responses.

\subsection{Coarse-to-Fine Group Policy Optimization}
\label{sec:gopo_objective}

We integrate the trajectory-level and turn-level advantages into a multiscale group-relative advantage for policy optimization. 
For the response $\boldsymbol{a}_{i,t}$ in trajectory $\boldsymbol{\tau}_i$, the final \methodname{} advantage is defined as,
\begin{equation}
A^{\mathrm\methodname{}}(\boldsymbol{a}_{i,t})
=
A^{\mathrm{E}}(\boldsymbol{\tau}_i)
+
\alpha(k) A^{\mathrm{S}}(\boldsymbol{a}_{i,t}),
\end{equation}
\begin{equation}
\mathrm{with}~~\alpha(k)=\mathtt{LinearDecay}(k;\alpha_{\rm init},\alpha_{\rm final}),
\label{eq:gopo_advantage}
\end{equation}
where $\alpha(k)$ denotes the decay weight with training step $k$ for balancing the two level signals. We set the turn-level coefficient smaller than the trajectory-level coefficient, i.e., $0\leq \alpha_{\rm final}\leq \alpha(k)< \alpha_{\rm init}\leq1$. This design keeps trajectory-level planning quality as the dominant signal and uses response length signals only as a refinement. Although the response constraint is applied only to successful trajectories, an overly large $\alpha(k)$ experimentally over-penalize response length, thereby hurting task performance. Please see the ablation study for details. Therefore, we gradually decay $\alpha(k)$ during training to reduce the strength of response-length normalization as the policy becomes more capable.

Finally, \methodname{} optimizes the policy with the same clipped group-relative objective backbone as GRPO-style methods:
\begin{equation}
\begin{aligned}
\mathcal{J}^{\mathrm\methodname{}}(\theta)
&=
\mathbb{E}_{\substack{
\boldsymbol{x}\sim p(X),\\
\{\boldsymbol{\tau}_i\}_{i=1}^{N}\sim\pi_{\theta_{\rm old}}
}}
\Bigg[
\frac{1}{\sum_{i=1}^{N}T_i}
\sum_{i=1}^{N}
\sum_{t=1}^{T_i}
\ell_{i,t} \\
&-
\beta
\mathbb{D}_{\mathrm{KL}}
\bigl(
\pi_{\theta}(\cdot\mid\boldsymbol{x})
\|
\pi_{\rm ref}(\cdot\mid\boldsymbol{x})
\bigr)
\Bigg],
\end{aligned}
\label{eq:gopo_objective}
\end{equation}
\begin{equation}
\begin{aligned}
\ell_{i,t}
=
\min\Big(
&\rho_{\theta}(\boldsymbol{a}_{i,t})
A^{\mathrm\methodname{}}(\boldsymbol{a}_{i,t}),\\
&\operatorname{clip}(\rho_{\theta}(\boldsymbol{a}_{i,t}),1-\epsilon,1+\epsilon)
A^{\mathrm\methodname{}}(\boldsymbol{a}_{i,t})
\Big).
\end{aligned}
\label{eq:clipped_surrogate}
\end{equation}

Here, $\rho_{\theta}(\boldsymbol{a}_{i,t})=
\pi_{\theta}(\boldsymbol{a}_{i,t}\mid\boldsymbol{s}_{i,t},\boldsymbol{x})/
\pi_{\theta_{\rm old}}(\boldsymbol{a}_{i,t}\mid\boldsymbol{s}_{i,t},\boldsymbol{x})$
is the importance sampling ratio, $\epsilon$ is the clipping coefficient, and $\beta$ controls the strength of KL regularization with respect to the reference policy $\pi_{\rm ref}$.

\begin{table*}[t]
\centering
\resizebox{\textwidth}{!}{
\begin{tabular}{llccccccc|cc}
\toprule
\multirow{2}{*}{Type} & \multirow{2}{*}{Method} & \multicolumn{7}{c|}{\textbf{ALFWorld}} & \multicolumn{2}{c}{\textbf{WebShop}} \\
 & & Pick & Look & Clean & Heat & Cool & Pick2 & All & Score & Succ.\\
\midrule
\multicolumn{10}{l}{\textit{Closed-Source Model}} \\
Prompting& GPT-4o & 75.3 & 60.8 & 31.2 & 56.7 & 21.6 & 49.8 & 48.0& 31.8 & 23.7\\
Prompting& Gemini-2.5-Pro & 92.8 & 63.3 & 62.1 & 69.0 & 26.6 & 58.7 & 60.3& 42.5 & 35.9\\
\midrule
\multicolumn{9}{l}{\textit{Qwen2.5-1.5B-Instruct}} \\
Prompting& ReAct & 17.4 & 20.5 & 15.7 & 6.2 & 7.7 & 2.0 & 12.8& 40.1& 11.3\\
Prompting& Reflexion & 35.3 & 22.2 & 21.7 & 13.6 & 19.4 & 3.7 & 21.8 & 55.8& 21.9\\
RL Training& PPO (with critic) & 64.8\textsubscript{\textpm3.5} & 40.5\textsubscript{\textpm6.9} & 57.1\textsubscript{\textpm4.9} & 60.6\textsubscript{\textpm6.6} & 46.4\textsubscript{\textpm4.0} & 47.4\textsubscript{\textpm1.9} & 54.4\textsubscript{\textpm3.1}& 73.8\textsubscript{\textpm3.0} & 51.5\textsubscript{\textpm2.9} \\
RL Training& RLOO & 88.3\textsubscript{\textpm3.0} & 52.8\textsubscript{\textpm8.6} & 71.0\textsubscript{\textpm5.9} & 62.8\textsubscript{\textpm8.7} & 66.4\textsubscript{\textpm5.5} & 56.9\textsubscript{\textpm4.7} & 69.7\textsubscript{\textpm2.5}& 73.9\textsubscript{\textpm5.6}& 52.1\textsubscript{\textpm6.7}\\
RL Training& GRPO & 85.3\textsubscript{\textpm1.5} & 53.7\textsubscript{\textpm8.0} & 84.5\textsubscript{\textpm6.8} & 78.2\textsubscript{\textpm7.9} & 59.7\textsubscript{\textpm5.0} & 53.5\textsubscript{\textpm5.6} & 72.8\textsubscript{\textpm3.6}& 75.8\textsubscript{\textpm3.5} & 56.8\textsubscript{\textpm3.8}\\
RL Training& EMPG & 85.5 & 33.5 & 78.9 & 76.2 & 74.7 & 89.1 & 73.7 & 80.4 & 60.8\\
RL Training& GiGPO\textsubscript{w/ std} & 94.4\textsubscript{\textpm5.9} & 67.5\textsubscript{\textpm4.6} & {94.8}\textsubscript{\textpm3.8} & \textbf{94.4}\textsubscript{\textpm7.8} & 79.8\textsubscript{\textpm4.7} & 76.4\textsubscript{\textpm5.4} & {86.7}\textsubscript{\textpm1.7} & 83.1\textsubscript{\textpm1.6}& 65.0\textsubscript{\textpm3.2}\\
RL Training& GiGPO\textsubscript{w/o std} & {96.0}\textsubscript{\textpm1.4} & {76.5}\textsubscript{\textpm3.9} & 91.8\textsubscript{\textpm5.5} & 91.3\textsubscript{\textpm6.3} & 71.7\textsubscript{\textpm8.4} & {79.5}\textsubscript{\textpm7.7} & 86.1\textsubscript{\textpm4.7} & 83.5\textsubscript{\textpm1.8} & {67.4}\textsubscript{\textpm4.5}\\
\midrule
\rowcolor{gray!20}RL Training& \textbf{\methodname{}} & \textbf{98.2}\textsubscript{\textpm1.1} & \textbf{85.1}\textsubscript{\textpm4.6} & \textbf{94.6}\textsubscript{\textpm4.7} & 93.8\textsubscript{\textpm6.0} & \textbf{82.4}\textsubscript{\textpm6.7} & \textbf{83.7}\textsubscript{\textpm8.0} & \textbf{91.3}\textsubscript{\textpm4.1} & \textbf{86.8}\textsubscript{\textpm1.5} & \textbf{77.2}\textsubscript{\textpm5.6}\\
\midrule
\multicolumn{9}{l}{\textit{Qwen2.5-7B-Instruct}} \\
Prompting& ReAct & 48.5 & 35.4 & 34.3 & 13.2 & 18.2 & 17.6 & 31.2 & 46.2 & 19.5\\
Prompting& Reflexion & 62.0 & 41.6 & 44.9 & 30.9 & 36.3 & 23.8 & 42.7& 58.1& 28.8\\
RL Training& PPO (with critic) & 92.3\textsubscript{\textpm4.0} & 64.0\textsubscript{\textpm8.4} & 92.5\textsubscript{\textpm2.4} & 89.5\textsubscript{\textpm7.0} & 80.3\textsubscript{\textpm2.0} & 68.8\textsubscript{\textpm8.3} & 80.4\textsubscript{\textpm2.7} & 81.4\textsubscript{\textpm3.1}& 68.7\textsubscript{\textpm5.1}\\
RL Training& RLOO & 87.6\textsubscript{\textpm4.3} & 78.2\textsubscript{\textpm8.3} & 87.3\textsubscript{\textpm5.8} & 81.3\textsubscript{\textpm7.6} & 71.9\textsubscript{\textpm5.2} & 48.9\textsubscript{\textpm8.4} & 75.5\textsubscript{\textpm4.6} & 80.3\textsubscript{\textpm3.2} & 65.7\textsubscript{\textpm4.0}\\
RL Training& GRPO & 90.8\textsubscript{\textpm5.1} & 66.1\textsubscript{\textpm6.7} & 89.3\textsubscript{\textpm5.4} & 74.7\textsubscript{\textpm6.9} & 72.5\textsubscript{\textpm5.4} & 64.7\textsubscript{\textpm7.3} & 77.6\textsubscript{\textpm5.2} & 79.3\textsubscript{\textpm2.8} & 66.1\textsubscript{\textpm3.7}\\ 

RL Training& EMPG & 92.9 & 75.2 & 74.8 & 86.3 & 73.7 & 65.3 & 78.5 & 81.0 & 69.3\\ 

RL Training& {GiGPO\textsubscript{w/ std}} & {97.7}\textsubscript{\textpm1.6} & 82.7\textsubscript{\textpm7.9} & \textbf{98.8}\textsubscript{\textpm1.6} & 83.7\textsubscript{\textpm7.2} & \textbf{89.3}\textsubscript{\textpm8.2} & 79.2\textsubscript{\textpm6.6} & {90.8}\textsubscript{\textpm1.3} & 84.4\textsubscript{\textpm2.9} & 72.8\textsubscript{\textpm3.2}\\
RL Training& {GiGPO\textsubscript{w/o std}} & 91.8\textsubscript{\textpm5.4} & {88.6}\textsubscript{\textpm6.3} & 95.9\textsubscript{\textpm3.2} & {90.2}\textsubscript{\textpm2.6} & 86.5\textsubscript{\textpm5.5} & {85.2}\textsubscript{\textpm7.5} & 90.2\textsubscript{\textpm2.3} & {86.2}\textsubscript{\textpm2.6} & \textbf{75.2}\textsubscript{\textpm3.8}\\
\midrule
\rowcolor{gray!20}RL Training& \textbf{\methodname{}} & \textbf{100.0}\textsubscript{\textpm0.0} & \textbf{92.1}\textsubscript{\textpm6.3} & 96.8\textsubscript{\textpm2.5} & \textbf{99.0}\textsubscript{\textpm1.4} & 88.3\textsubscript{\textpm6.8} & \textbf{89.7}\textsubscript{\textpm7.8} & \textbf{94.4}\textsubscript{\textpm2.1} & \textbf{88.5}\textsubscript{\textpm3.2} & \textbf{80.6}\textsubscript{\textpm4.7}\\
\bottomrule
\end{tabular}
}
\caption{Evaluation results on ALFWorld and WebShop. For each RL training method, we report the mean and variance over three random seeds. The ALFWorld contains six categories: Pick \& Place (Pick), Examine in Light (Look), Clean \& Place (Clean), Heat \& Place (Heat), Cool \& Place (Cool), and Pick Two \& Place (Pick2). Most entries in this table are reported by Feng et al. ~\cite{feng2026group}.  Notably, the baseline GiGPO\textsubscript{w/ std} denotes $F_{\text{norm}} = \text{std}$, and GiGPO\textsubscript{w/o std} denotes $F_{\text{norm}} = 1$.}
\vspace{-10pt}
\label{tab:main}
\end{table*}

\begin{table*}[th]\small
\centering
\resizebox{\textwidth}{!}{%
\begin{tabular}{llcccccc}
\toprule
\textbf{Type} &
\textbf{Model} &
\textbf{Measure} &
\textbf{Test-Cond.} &
\textbf{Find} &
\textbf{Chem-Mix} &
\textbf{Lifespan} &
\textbf{Overall} \\
\midrule

\multirow{6}{*}{Proprietary}
& {GPT-4o} & 15.09 & 6.02 & 38.64 & {20.00} & 73.33 & 21.00 \\
& {Qwen-Max} & 9.43 & 0.00 & 34.09 & {20.00} & 40.00 & 13.50 \\
& {Gemini-2.5-Flash} & 11.32 & 0.00 & {54.55} & 0.00 & {80.00} & 21.00 \\
& {OpenAI o4-mini} & {20.75} & {14.46} & 47.73 & 0.00 & \textbf{100.00} & {29.50} \\
& {OpenAI o3} & \textbf{47.17} & {25.30} & {56.82} & \textbf{40.00} & 66.67 & {41.50} \\
& {Gemini-2.5-Pro} & 9.43 & 0.00 & 29.55 & 0.00 & 46.67 & 12.50 \\

\midrule

\multirow{3}{*}{Models $\ge$100B}
& {Qwen3-235B-A22B} & {11.32} & {4.82} & {59.09} & {20.00} & {66.67} & {23.50} \\
& {DeepSeek-V3-0324} & 0.00 & 0.00 & 2.27 & 0.00 & 0.00 & 0.50 \\
& {DeepSeek-R1-0528} & {1.89} & 0.00 & {11.36} & 0.00 & {20.00} & {4.50} \\

\midrule

\multirow{3}{*}{Models $<$100B}
& {Qwen2.5-72B-Instruct} & 7.55 & 1.20 & 15.91 & {20.00} & 40.00 & 9.50 \\
& {Qwen3-32B} & 5.66 & 1.20 & 31.82 & 0.00 & 66.67 & 14.00 \\
& {Llama-3.1-70B-Instruct} & {24.53} & 4.82 & 40.91 & \textbf{40.00} & {86.67} & 25.00 \\
\midrule
\multirow{4}{*}{RL Training}
& {AgentGym-RL-3B} & 20.75 & 28.92 & 0.00 & 0.00 & 66.67 & 22.50 \\
& {AgentGym-RL-7B} & 24.53 & {59.04} & 65.91 & 0.00 & 66.67 & 50.50 \\
& {ScalingInter-7B} & {33.96} & 55.42 & {88.64} & 0.00 & {73.33} & {57.00} \\
\rowcolor{gray!20}& \textbf{\methodname{}-7B} & {40.52} & \textbf{69.38} & \textbf{91.67} & 0.00 & {85.94} & \textbf{68.46} \\
\bottomrule
\end{tabular}
}
\caption{Evaluation results on SciWorld over three random seeds. The task Test-Cond. means test conductivity and Chem-Mix means chemistry mix. Baseline entries are reported by Zhang et al. ~\cite{zhang2026agentv}.}
\label{tab:main2}
\vspace{-0pt}
\end{table*}

\begin{figure*}[t]
   \begin{center}
   \includegraphics[width=0.99\linewidth]{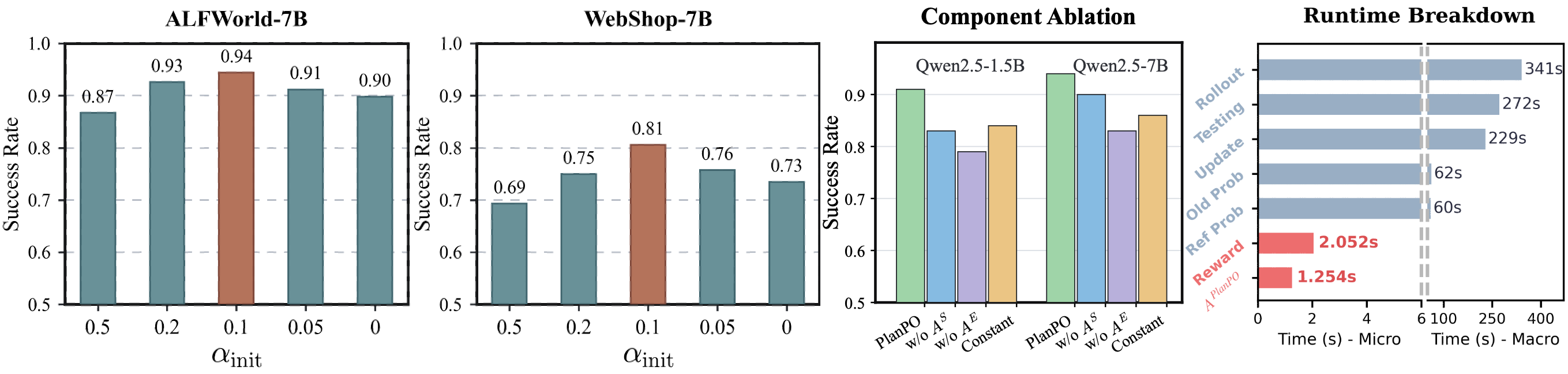}
   \end{center}
   \vspace{-0.1in}
   \caption{\textbf{Left two panels:} Ablation study on the weight schedule $\alpha(k)$ with $\alpha_{\mathrm{init}}=2\alpha_{\mathrm{final}}$ in PlanPO. \textbf{Third panel:}  Component ablation. We compared the ablation results of \textit{PlanPO with default decay}, \textit{trajectory-level advantage $A^E$ only}, \textit{turn-level advantage $A^S$ only}, and \textit{PlanPO with constant $\alpha=0.1$}. \textbf{Right panel:} Runtime breakdown of each PlanPO iteration on four NVIDIA A40 GPUs. Compared to the GRPO, the additional computation introduced by \methodname{} lies in the reward and advantage stages, which is negligible. Conversely, \methodname{} can reduce overall task runtime by 12.5\%}
   \label{fig:ablation}
   \vspace{-10pt}
\end{figure*}


\begin{theorem}[Bias--variance trade-off in PlanPO]
Consider a fixed response $a_{i,t}$ at optimization step $k$, with the randomness induced by repeated same-task group sampling. Let $A^{E,\star}=\mathbb{E}_{\mathcal G}[A^E]$ and $A^{S,\star}=\mathbb{E}_{\mathcal G}[A^S]$ denote the expected trajectory-level signal and turn-level refinement, respectively. Assume that their centered estimation errors have variances $v_E$ and $v_S$ and non-negative covariance $c$. Relative to the full-refinement target $A^\star=A^{E,\star}+A^{S,\star}$, the PlanPO estimator $A_\alpha=A^E+\alpha A^S$, where $\alpha\in[0,1]$, satisfies
\[
\begin{aligned}
\operatorname{Bias}^2(A_\alpha)
&=
(1-\alpha)^2(A^{S,\star})^2,\\
\operatorname{Var}(A_\alpha)
&=
v_E+\alpha^2v_S+2\alpha c.
\end{aligned}
\]
Consequently, increasing $\alpha$ decreases the squared bias while increasing the variance, with:
\[
\begin{aligned}
0
&\leq
\operatorname{Bias}^2(A_\alpha)
\leq
(A^{S,\star})^2,\\
v_E
&\leq
\operatorname{Var}(A_\alpha)
\leq
\left(\sqrt{v_E}+\sqrt{v_S}\right)^2.
\end{aligned}
\]
\end{theorem}

This theorem shows that $\alpha(k)$ relates to PlanPO's bias--variance trade-off. A larger $\alpha(k)$ preserves more turn-level refinement but raises sampling variance, whereas a smaller $\alpha(k)$ reduces variance at the cost of greater shrinkage bias. The technical supplement provides detailed proof.


\section{Experiments} 
We evaluate PlanPO across a range of multi-turn environments. Specifically, the experiments aim to answer three questions: (1) \textit{How well does PlanPO perform overall across these challenging environments?} (2) \textit{How does PlanPO achieve its performance gains?} (3) \textit{Does PlanPO acquire the capabilities of task generalization and planning awareness  rather than overfitting to task-specific success patterns?}

\subsection{Experimental Setup} 
\paragraph{Benchmarks.} We train and evaluate PlanPO on three challenging multi-turn benchmarks: ALFWorld \cite{shridhar2020alfworld}, WebShop \cite{yao2022webshop}, and SciWorld \cite{wang2022scienceworld}. ALFWorld is an embodied household environment for evaluating long-horizon textual reasoning and decision-making. In each episode, the agent receives a concrete task instruction sampled from $3{,}827$ tasks from six categories. WebShop is an interactive web-shopping environment containing nearly $1.1$ million products and $12{,}000$ user instructions. SciWorld is designed for scientific tasks and provides APIs through which agents can manipulate scientific instruments and conduct experiments. The technical supplement provides further descriptions of the benchmarks. 

\paragraph{Baselines.} For ALFWorld and WebShop, we compare PlanPO against three categories of competitive baselines: (1) Proprietary models: GPT-4o \cite{achiam2023gpt} and Gemini-2.5-Pro \cite{team2023gemini}; (2) training-free prompting agents: ReAct \cite{yao2022react} and Reflexion \cite{shinn2023reflexion}; and (3) RL-based methods: PPO \cite{schulman2017proximal}, RLOO \cite{ahmadian2024back}, GRPO \cite{shao2024deepseekmath}, EMPG \cite{wang2026harnessing}, and GiGPO \cite{feng2026group}, which cover representative actor--critic and group-relative optimization approaches. For SciWorld, we consider (1) training-free agents based on the OpenAI \cite{achiam2023gpt}, Gemini \cite{team2023gemini}, Qwen \cite{yang2025qwen3}, Llama \cite{grattafiori2024llama}, and DeepSeek \cite{liu2024deepseek,guo2025deepseek} model families, spanning both proprietary models and open-source models of different scales; and (2) RL-based methods: AgentGym-RL \cite{xi2026agentgymrl}, a GRPO framework for training across multiple environments, and ScalingInter \cite{xi2026agentgymrl}, a group-based RL method that reports strong performance on SciWorld. 

\paragraph{Implementation details.} We employ Qwen2.5-1.5B-Instruct and Qwen2.5-7B-Instruct as the base models for training experiments. To ensure fair comparisons, the hyperparameter settings of PlanPO follow the existing RL framework \cite{feng2026group,xi2026agentgymrl} in each benchmark. Specifically, we use a group size of $N=8$, a learning rate of $1\times10^{-6}$, and a KL-penalty coefficient of $1\times10^{-3}$ for SciWorld and $0.01$ for the other environments. The maximum number of interaction turns is set to $50$ for ALFWorld, $15$ for WebShop, and $20$ for SciWorld. We train for $150$ steps on ALFWorld and WebShop and for $200$ steps on SciWorld. On ALFWorld and WebShop, we set the turn-level weight $\alpha_{\mathrm{init}}=2\alpha_{\mathrm{final}}=0.1$ in PlanPO. For the exploratory SciWorld, we reduce it to $0.05$ without further tuning to improve training stability. Finally, all our experiments were run on 6 NVIDIA H200s and 8 NVIDIA A40s. The technical supplement provides additional implementation details.
\subsection{Experiment Results}

\paragraph{Performance on ALFWorld and WebShop.}
As shown in Table~\ref{tab:main}, PlanPO substantially outperforms advanced closed-source models on both benchmarks. For the open-source Qwen2.5-1.5B-Instruct and Qwen2.5-7B-Instruct models, RL methods are significantly stronger than prompting-based methods. Nevertheless, PlanPO also outperforms existing powerful baselines, e.g., GiGPO and EMPG. For example, on the 1.5B model, PlanPO achieves an overall success rate of $91.3\%$ across the six ALFWorld categories, improving over GRPO by $+18.5$ points, while requiring only several regularization operations with negligible computational cost. 

\paragraph{Performance on SciWorld.}
In Table~\ref{tab:main2}, PlanPO achieves consistently the best overall scores on complex and exploratory scientific scenarios, notably improving the GRPO-based AgentGym-RL from $50.5$ to $68.46$. However, we observe that RL training methods exhibit a consistent failure pattern in the Chem-Mix domain, which may be due to the model's limited understanding and exploration ability required for rigorous scientific analysis~\cite{xi2026agentgymrl}.

\subsection{In-depth Analysis}
\paragraph{Ablation Study.}
To verify how PlanPO achieves performance gains, we performed two ablation studies on the scheduling coefficient $\alpha(k)$ and each component. As shown in left Figure ~\ref{fig:ablation}, we compared a series of initial $\alpha_{\mathrm{init}}=\{0.5,0.2,0.1,0.05,0\}$. We observed that $\alpha_{\mathrm{init}}=0.1$ achieved the best performance, while both smaller values (e.g., $\alpha_{\mathrm{init}}=0$) and excessively large values (e.g., $0.5$) degraded performance.  This matches the theoretical bias--variance trade-off, i.e., the turn-level signal is useful as a refinement, but should not dominate the trajectory-level planning signal. In the third panel of Figure \ref{fig:ablation}, we observe that ablating any level signal significantly degrades PlanPO's performance, especially without $A^E$. Regarding turn-level advantages, PlanPO with coefficient decay settings consistently outperforms those using constant coefficients. In summary, its performance gains primarily stem from the $A^E$, while a properly weighted $A^S$ further enables optimal performance.


\begin{table}[t]
\centering
\begin{tabular}{lccc}
\toprule
\textbf{Type} & \textbf{Method} & \textbf{In-Success} & \textbf{Out-Success} \\
\midrule
Prompting & GPT-4o & 48.0 & 46.0 \\
Prompting & Gemini2.5 & 60.3 & 50.5 \\
\midrule
RL Training & PPO & 54.4\textsubscript{\textpm3.1} & 50.9\textsubscript{\textpm7.6} \\
RL Training & RLOO & 69.7\textsubscript{\textpm2.5} & 68.7\textsubscript{\textpm10.7} \\
RL Training & GRPO & 72.8\textsubscript{\textpm3.6} & 70.1\textsubscript{\textpm2.5} \\
RL Training & GiGPO & \textbf{86.7\textsubscript{\textpm1.7}} & \textbf{82.4\textsubscript{\textpm2.0}} \\
\midrule
\rowcolor{gray!20}
\textbf{RL Training} & \textbf{PlanPO} & 91.3\textsubscript{\textpm4.1} & 87.1\textsubscript{\textpm3.6} \\
\bottomrule
\end{tabular}
\caption{Generalization evaluation with three seeds in ALFWorld using Qwen2.5-1.5B-Instruct. Some entries are reported by He et al. \cite{he2026hierarchy}. The In-Success reports the success rate on the in-distribution tasks, while the Out-Success reports the success rate on out-of-distribution tasks.}
\label{tab:ood}
\vspace{-10pt}
\end{table}

\paragraph{Generalization Verification.}
The gains of \methodname{} remain strong when the evaluation tasks differ from the training distribution. As shown in Table~\ref{tab:ood}, \methodname{} achieves $87.1\%$ success on out-of-distribution ALFWorld tasks, improving over GRPO by $+17.0$ points and over GiGPO by $+4.7$ points. The gap between in-distribution and out-of-distribution performance is also modest, decreasing from $91.3\%$ to $87.1\%$. This suggests that group-based \methodname{} does not merely memorize some better planning templates, but learns planning-aware abilities that generalize to unseen task configurations.

\begin{figure}[b]
   \vspace{-10pt}
   \begin{center}
   \includegraphics[width=0.95\linewidth]{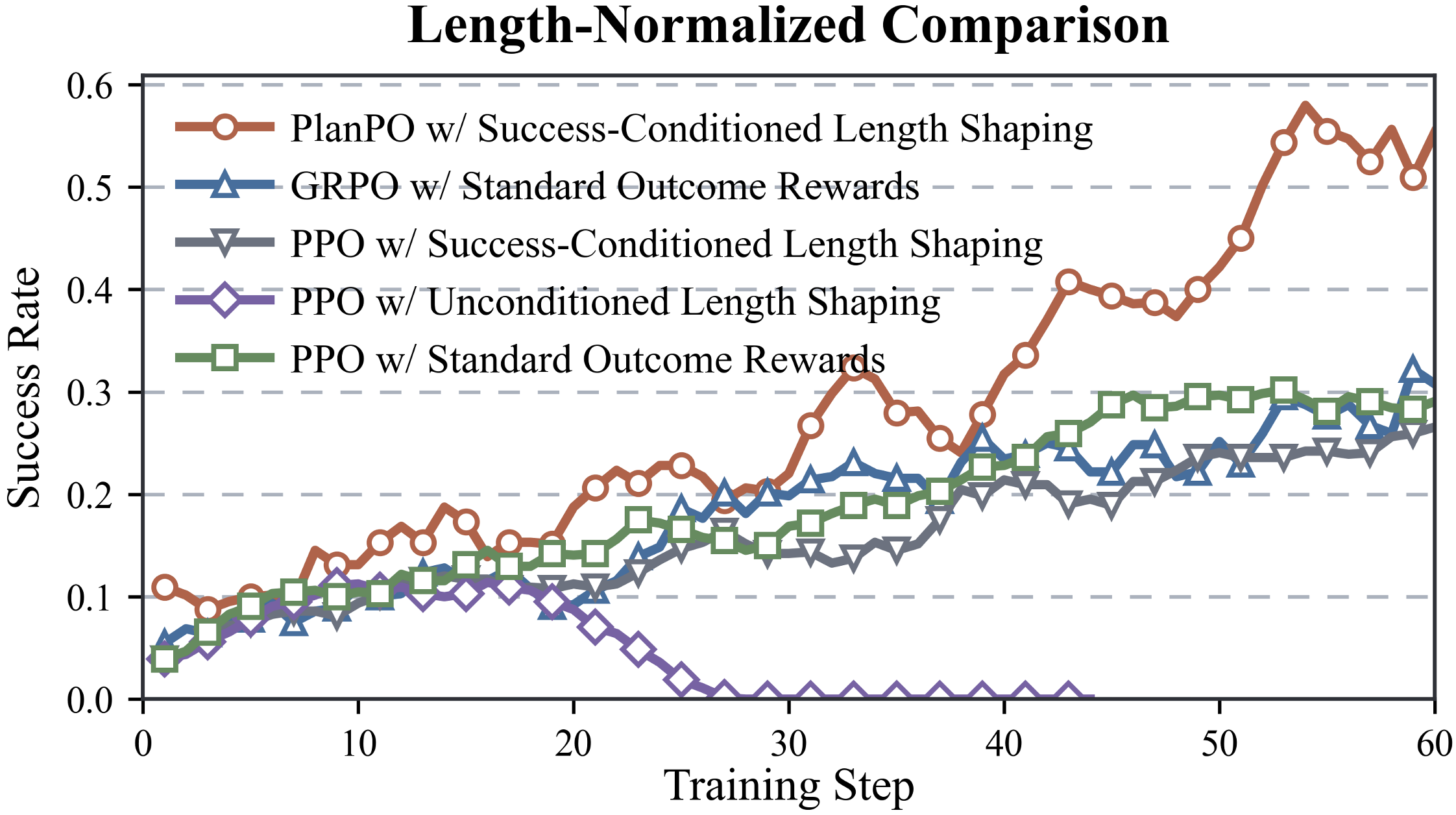}
   \end{center}
   \vspace{-0.1in}
   \caption{Length-normalized Comparison Analysis with different reward shaping settings in ALFWorld.}
   \label{fig:analysis}
   \vspace{-15pt}
\end{figure}

\begin{figure}[th]
   \begin{center}
   \includegraphics[width=1.0\linewidth]{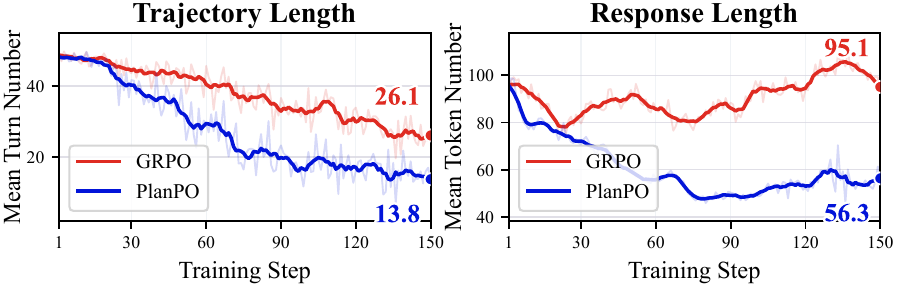}
   \end{center}
   \vspace{-0.1in}
   \caption{Mean Length Comparison in ALFWorld tasks.}
   \label{fig:planning}
   \vspace{-15pt}
\end{figure}

\paragraph{Length-Normalized Method Comparison.}
Figure~\ref{fig:analysis} further isolates the role of success-conditioned length normalization. Directly applying unconditional length shaping in PPO quickly collapses to nearly zero success, showing that shorter generations or trajectories are not intrinsically better. Success conditioning alone is also insufficient, as PPO with success-conditioned length shaping remains close to standard PPO. By contrast, \methodname{} steadily improves after training begins and reaches around $0.55$ success near the end, clearly above GRPO and PPO variants. These results indicate that the key benefit comes from coupling length-induced signals with group-relative policy optimization, which turns heterogeneous successful rollouts into useful supervision. Note that the reward shaping operation and outcome reward size $R(\tau)=10$ involved are consistent with PlanPO settings, except for the condition control and basic algorithm.

\paragraph{Planning-Aware Strategies.} We further examine whether \methodname{} induces planning-aware behavior beyond improving final task success. As shown in Figure~\ref{fig:planning}, \methodname{} consistently produces shorter trajectories than GRPO during training, reducing the mean number of interaction turns to $13.8$ compared with $26.1$ for GRPO. This suggests that \methodname{} learns to reach task goals through more direct interaction paths, avoiding redundant transitions, repeated trials, and unnecessary detours. At the response level, \methodname{} also yields more compact generations, with the mean response length decreasing to $56.3$ tokens, while GRPO remains substantially more verbose at $95.1$ tokens. Importantly, this behavior does not arise from naive length minimization: as shown in Figure~\ref{fig:analysis}, unconditional length shaping hurts task success, whereas \methodname{} applies length normalization only under success-conditioned group-relative comparisons. These results indicate that \methodname{} encourages planning-aware strategies by preferring successful rollouts that are both interaction-efficient and response-concise, thereby turning heterogeneity among successful trajectories into a useful signal for learning generalizable agentic planning abilities.



\section{Related Work}

RL has become a central recipe for improving LLM reasoning and interactive agents \cite{sheng2025hybridflow,shao2024deepseekmath,zeng2026glm,kimiteam2026kimik3openfrontier}. PPO-based RLHF \cite{schulman2017proximal} trains a value model to reduce variance, while lighter critic-free alternatives such as ReMax \cite{li2023remax}, RLOO \cite{ahmadian2024back}, and GRPO \cite{shao2024deepseekmath} replace the learned critic with sampled baselines or group-relative rewards, improving scalability for LLMs. Recent GRPO variants further refine this objective \cite{liu2025learn,cui2025entropy,zheng2025group,lin2026cppo}. DAPO improves stability with decoupled clipping, dynamic sampling, and overlong-response shaping \cite{yu2026dapo}, Dr. GRPO analyzes length and reward-variance normalization biases \cite{liu2025understanding}, and GMPO stabilizes updates by changing the aggregation of token-level rewards \cite{zhao2025geometric}. In parallel, agentic RL methods extend outcome-reward optimization to long-horizon interaction, including WebSailor \cite{li2025websailor}, GiGPO \cite{feng2026group}, HGPO \cite{he2026hierarchy}, and A2TGPO \cite{chen20262}, which study credit assignment through episode-, state-, history-, or turn-grouped advantages. Different from these methods, our work focuses on planning-aware credit assignment under successful-rollout heterogeneity.

\section{Conclusion}
This paper addresses successful-rollout heterogeneity in group-relative RL for multi-turn agentic LLMs. We propose \methodname{}, which constructs success-conditioned length-induced advantages from trajectory-level interaction and token-level response generation. The experiments across multiple benchmarks demonstrate that \methodname{} consistently outperforms existing methods and achieves the best task performance with stronger generalization, planful Strategies, and negligible overhead. Our results suggest that successful-rollout heterogeneity provides a simple and scalable signal for learning planning-aware and generalizable behaviors.

\bibliography{aaai2027}
\end{document}